\documentclass[letterpaper, 10 pt, journal, twoside]{ieeetran}  % Use this line for a4 paper

\IEEEoverridecommandlockouts                              % This command is only needed if 
\usepackage[nolist,nohyperlinks]{acronym}
\usepackage{amsmath} % assumes amsmath package installed
\usepackage{cases}
\usepackage{amssymb}  % assumes amsmath package installed
\usepackage[colorinlistoftodos]{todonotes}
\usepackage{hyperref}
\usepackage{multirow}
\usepackage{booktabs}
\usepackage[colorinlistoftodos]{todonotes}
\usepackage{textcomp}
\usepackage{gensymb}
\usepackage[separate-uncertainty=true]{siunitx}
\usepackage{url}
\usepackage{mathtools}
\usepackage{subcaption}
\usepackage{bm}  
\usepackage{soul}
\usepackage{dblfloatfix}
\usepackage{cite}

\providecommand{\keywords}[1]{\textbf{\textit{Index terms---}} #1}

\title{Deep Measurement Updates for Bayes Filters}

\author{Johannes Pankert, Maria Vittoria Minniti, Lorenz Wellhausen, Marco Hutter % <-this % stops a space
\thanks{This research was supported by the Swiss Federal Railways (SBB) and the Swiss National Science Foundation through the National Centre of Competence in Digital Fabrication (NCCR dfab).}
\thanks{All authors are with the Robotic Systems Lab, ETH Zurich. {\tt\small \{pankert|mminniti|lorenwel|hutter\}@ethz.ch}
}% <-this % stops a space
}

\begin{document}
%These are last resort solutions to reduce the amount of free space after the figures 
%\setlength{\textfloatsep}{10pt}
\setlength{\dbltextfloatsep}{15pt}

\maketitle
%\thispagestyle{empty}
%\pagestyle{empty}

%\markboth{IEEE Robotics and Automation Letters. Preprint Version. Accepted October, 2021}
%{Pankert \MakeLowercase{\textit{et al.}}: Deep Measurement Updates} 

%%%%%%%%%%%%%%%%%%%%%%%%%%%%%%%%%%%%%%%%%%%%%%%%%%%%%%%%%%%%%%%%%%%%%%%%%%%%%%%%
\begin{abstract}
Measurement update rules for Bayes filters often contain hand-crafted heuristics to compute observation probabilities for high-dimensional sensor data, like images.
In this work, we propose the novel approach Deep Measurement Update (DMU) as a general update rule for a wide range of systems.
DMU has a conditional encoder-decoder neural network structure to process depth images as raw inputs. Even though the network is trained only on synthetic data, the model shows good performance at evaluation time on real-world data.\\
With our proposed training scheme \textit{primed data training}, we demonstrate how the DMU models can be trained efficiently to be sensitive to condition variables without having to rely on a stochastic information bottleneck.
We validate the proposed methods in multiple scenarios of increasing complexity, beginning with the pose estimation of a single object to the joint estimation of the pose and the internal state of an articulated system.
Moreover, we provide a benchmark against Articulated Signed Distance Functions(A-SDF) on the RBO dataset as a baseline comparison for articulation state estimation.
%We present a baseline comparison in articulation state estimation to Articulated Signed Distance Functions (A-SDF) on the RBO dataset, where DMU achieves similar accuracy while being 1000x faster in inference.
%Compared to the baseline Articulated Signed Distance Functions (A-SDF), DMU achieves similar accuracy in articulation state estimation while being 1000x faster in inference.
%In this work, we investigate the use of deep learning methods to compute measurement updates for Bayes filters. The goal is to infer the conditional probability of a measurement occurring given the system is in a certain state.
%We consider depth images as measurements and show how the likelihood of a system state can be inferred while being robust to occlusions.
%We validate the proposed methods in multiple scenarios of increasing complexity, beginning with the pose estimation of a single object to the estimation of the internal state of an articulated system.
\end{abstract}
\section{INTRODUCTION}
\begin{figure}[t]
\centering
\includegraphics[width=\columnwidth]{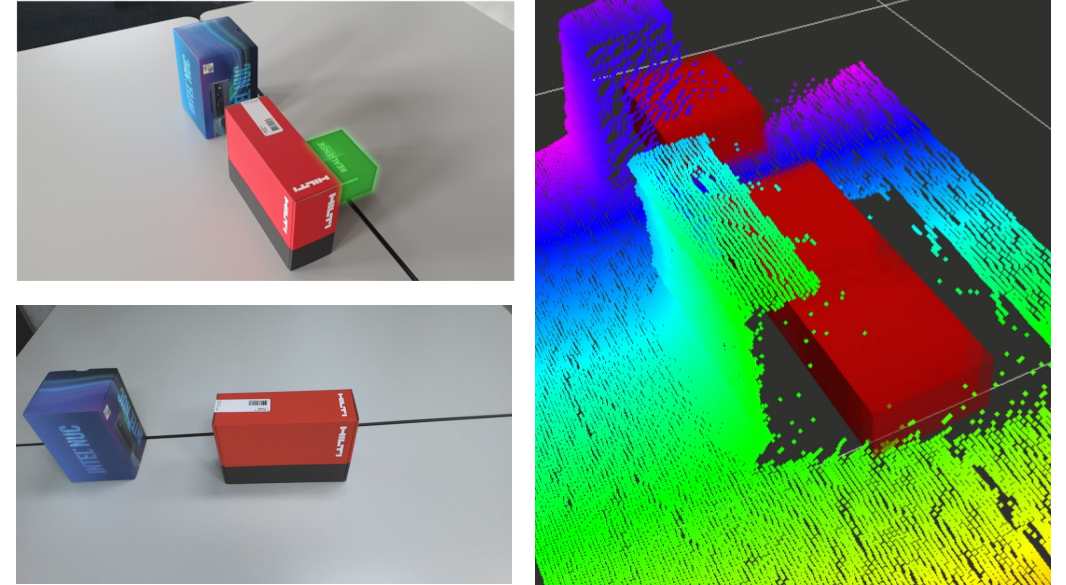}
\caption{Upper Left: A scene with three boxes. The pose of the small box highlighted in green should be estimated. \\
Lower left: The same scene from the point of view of the depth sensor. The target box is occluded. \\
Right: Particle filter estimation of the box pose.
A depth image of the scene is displayed as a point cloud. The distribution of 100 particles is visualized with overlaying red semi-transparent boxes. DMU is run on the depth image and assigns a higher observation probability to the particles behind the obstacles than those in visible free space.}
\label{fig:teaser}
\end{figure}

\IEEEPARstart{I}{n} computer vision, many research works have focused on the problem of directly inferring state information from sensor measurements \cite{redmonYouOnlyLook2016, heMaskRCNN2018, xiang2017posecnn}.
However, only partial observations are available for many real-world robotics applications, and the entire system state cannot be inferred from a single measurement.
In Fig.~\ref{fig:teaser}, an example is presented in which a target object is fully occluded. Free space can however be observed and information on the object pose can be indirectly inferred.
A sequence of measurements and prior knowledge is often needed to estimate the full system state.
\\
Bayes Filters provide a general way to solve this problem \cite{thrun2005probabilistic}. Using the Markov assumption, this framework aims to recursively compute the belief $bel(x^{[t]})$ of the state $x$ at time $t$ given the prior belief $bel(x^{[t-1]})$ and the observation $y^{[t]}$.
In this work, we take a detailed look at how to update the predicted belief $\overline{bel(x^{[t]})}$ from a measurement:
\begin{align}
bel(x^{[t]}) &= \eta p(y^{[t]}|x^{[t]}) \overline{bel(x^{[t]})}.
\label{eq:particle_filter_update}
\end{align}
In \eqref{eq:particle_filter_update}, $\eta$ is a normalizing factor and $p(y^{[t]}|x^{[t]})$ is the conditional probability density function (CPDF) of a measurement $y^{[t]}$, conditioned on the current system state $x^{[t]}$.
In the following, we drop the $[t]$ superscripts since we do not focus on the time evolution aspect of Bayes filtering.\\
Finding the CPDF can be viewed as the inverse problem of inferring the state from measurements. In practical robotic applications, this conditional probability is difficult to determine.
Often, hand-crafted sensor processing pipelines are designed and tuned and measurement probabilities are being assigned solely based on heuristics \cite{thrun2005probabilistic, montemerloFastSLAMScalableMethod2007}.
In this work, we want to use deep learning to find a systematic approach for determining $p(y|x)$.We propose a conditional auto-encoder (CAE) architecture to learn the probability distribution of possible measurements that can be observed for our system, conditioned on the ground truth system state. Learning the probability distribution associated with high-dimensional training data is addressed by \textit{generative models}, which include several deep-learning architectures \cite{salakhutdinov2009deep, hinton2006fast, goodfellow2014generative, kingma2013auto, sohn2015learning, rezende2014stochastic}.
% In this work, we use a generative model to learn the probability distribution of possible measurements that can be observed for our system, conditioned on the ground truth system state.
% In general, Variational Auto Encoders (VAEs) can perform efficient probabilistic inference by encoding the input data into a Gaussian latent vector and approximately maximizing the evidence lower bound (ELBO).
Conditional Variational Auto Encoders (CVAEs) \cite{sohn2015learning} offer a general method to perform an approximate inference of the probability of the training data, based on some input observation. They have been shown to be successful in generating output samples that behave according to the underlying probability distribution of the training data. However, numerical evaluation of the marginalized conditional probability $p(y|x)$ requires Monte Carlo sampling on the latent code distribution \cite{sohn2015learning} for each queried system state $x$, making inference slow. Furthermore, balancing the evidence lower bound objective (ELBO) to condition the latent state distribution and the reconstruction loss adds more complexity to the task.\\
The proposed CAE architecture makes the sampling step superfluous. It allows to efficiently compute and evaluate the measurement probability for integration in the filter update rule.
Learning to extract a state-independent latent code from the input observation is an additional challenge. The difficulty increases when the state dimension is much smaller than the latent code dimension.
% the network dependency on the state becomes intractable if the state dimension is much smaller than the latent code dimension. 
For instance, in visual object localization, the state includes the object pose and the latent distribution needs to describe the, possibly highly complex, image background.
In preliminary tests, we found that the trained models relied entirely on the latent code to capture both background and state information. The CAE degenerated to a non-conditional autoencoder.\\
To overcome this issue, we propose a new training method, named \textit{primed data training}, which requires the network to synthesize observations with the same background information as the input, but a different conditional state.
\textit{Primed data training} successfully conditions the network on the state variable. We validate the training scheme on our proposed CAE architecture, although we point out its generality and possible applicability with different conditional models, such as CVAEs or Conditional Normalizing Flows \cite{winkler2019learning}.
% The approximation is computed by Monte Carlo sampling on the latent code distribution \cite{sohn2015learning}, \cite{rezende2014stochastic}.
% This method is only accurate if a very large number of samples is drawn which makes inference slow.
% Furthermore, the sampling may lead to non-smooth probability estimates, that would be unusable in the update rule \eqref{eq:particle_filter_update}.
% Lastly, balancing the evidence lower bound objective (ELBO) to condition the latent state distribution and the reconstruction loss has shown to be difficult. 

% In preliminary tests we found that the trained CVAE models would either fully rely on the latent code ignoring the condition variable or solely base the prediction on the condition variable and disregard the latent code.
% For those reasons, the application of a CVAE for the Bayes filter update rule is not straightforward.
% To overcome this issue, we designed a conditional autoencoder (CAE) architecture that makes the sampling steps of the CVAE superfluous. Instead a new training method is proposed that successfully achieves a conditioning on the state variable $x$ without the need for a KL-Divergence loss function.
\subsection{Related Work}
Methods for single-shot object pose detection are well established in the computer-vision and robotics literature \cite{xiang2017posecnn, zakharovDPOD6DPose2019a}. Those works are conceptually very different from ours since they solely focus on the specific problem of pose estimation and cannot easily be applied to a broader class of state estimation problems.
DeepIM \cite{liDeepIMDeepIterative2020} uses a \textit{render-and-compare} approach to refine 6D pose estimates. It uses synthetically rendered images that a learned model compares to the input image. In our work, we also render images and compare them to the input image. However, our rendering is done by a decoder network that can process both the state provided explicitly and information extracted from the input image by the encoder network.\\
One of the demo cases we present in this paper is articulation state estimation. We benchmark against Articulated Signed Distance Functions (A-SDF) \cite{mu2021sdf}, a method that shows good performance on the RBO dataset \cite{rbo2019}. Unlike our work, A-SDF requires pre-processing with segmentation masks and is prohibitively slow for real-time applications.\\
In recent years, Bayes Filtering has regained the attention of researchers from the deep learning community.
Methods like Deep Kalman Filter learn variational models from sequence data \cite{krishnanDeepKalmanFilters2015, beckerRecurrentKalmanNetworks2019, DBLP:conf/iclr/KarlSBS17}. In contrast to our work, their learned system states are latent variables that cannot be interpreted directly. \\
Particles filter networks \cite{karkusParticleFilterNetworks2018a} learn an observation model to compute the weights required for resampling in a particle filter. In contrast to our work, they do not use a generative model to predict a measurement given a particle state but directly infer weights.
This requires them to train the system end-to-end on sequence data since no ground truth weights are available.\\
Some works use learned models to extract state information from input images and process the information with Bayes filters \cite{avantRigidBodyDynamics2020}. This approach is different from ours since we do not require problem-specific heuristics to perform measurement updates.
\subsection{Contributions}
The contribution of this paper is twofold:
A new approach to learn measurement update rules for Bayes filters is presented.
%To our best knowledge, this is the first work that uses a deep neural network as a method to derive measurement update rules for Bayes filters.
Second, we propose a novel training scheme, \textit{primed data training}, which enforces decoder sensitivity to condition variables without introducing a stochastic information bottleneck usually found in CVAEs.\\
The learned models, which are trained only on synthetic data, are validated in real-world experiments.
% Since the output $z_t$ of the encoder part of the CVAE is forced to be unit-normally distributed during training, we can use this at inference time to compute the likelihood $p(z_t|x_t)$ of a generated latent vector:
% \begin{align}
% p(z_t|x_t) &\approx e^{-\frac{1}{2}(z_t^T z_t)}
% \end{align}
% Our hypothesis is that $p(y_t|x_t)$ and $p(z_t|x_t)$ are correlated\replaced{and that}{,} we can use this to update our belief state $bel(x_t)$.
% \todo[inline]{This sentence is unclear and needs to be refined}

%As a demonstration case, our problem consists in the estimation of the observation probability of a cube, whose position and orientation vary over a dataset of depth images. This scenario is interesting since $p(y_t|x_t)$ is multi-modal because of its symmetry properties, something that direct state inference methods do not explicitly handle~\cite{xiang2017posecnn}.

\section{METHODS}
\label{sec:methods}
%In the following section, we describe the general idea of the deep measurement update method. Afterward, a method for training the underlying networks with synthetic data is presented.
%In the experimental section, we will share our results on depth images as measurements. The underlying approach is however general and could transfer to other sensor modalities as well.
\subsection{Deep Measurement Update}
\label{ss:dmu}
\begin{figure*}[htp]
	\centering
	\includegraphics[width=\linewidth]{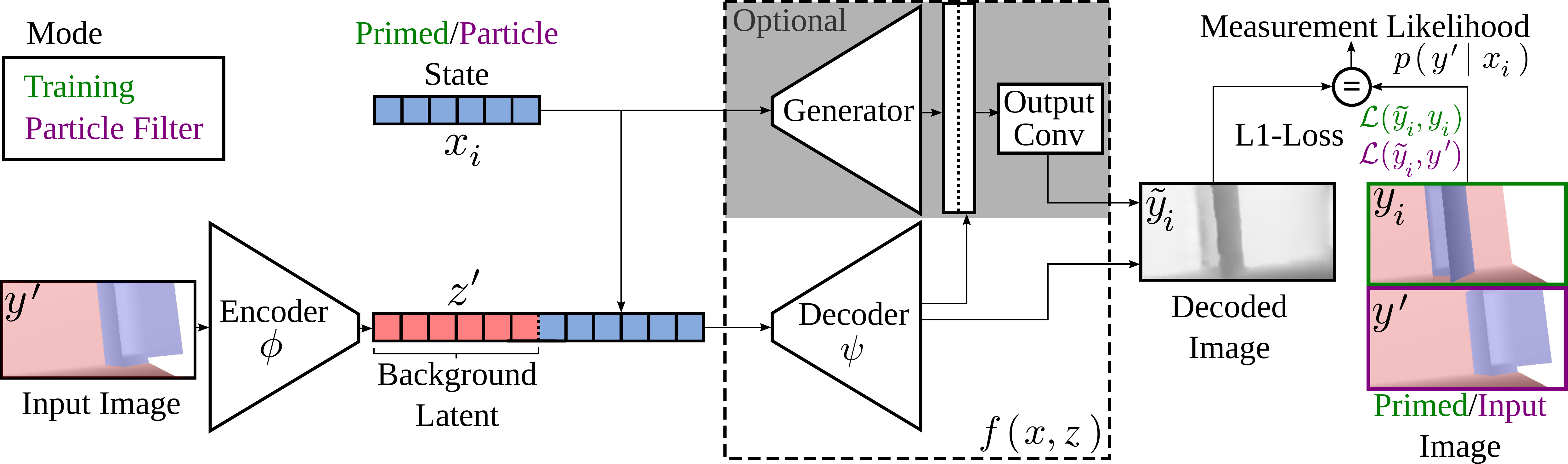}
	\caption{
% 	The CAE transcodes a depth image $y'$ with an unknown state $x'$ into a depth image $y_i$ given the particle state $x_i$. We infer the intractable similarity of $(x', x_i)$ by computing a loss function between the two depth images $(y', y_i)$.
	The encoder network compresses the background information of the input image, shaded in red, into a latent vector. The latent vector is augmented with the condition vector (i.e. the particle state of a Bayes filter, relevant image part shaded blue). The augmented latent is then decompressed by the decoder and fed to the output module. 
	We also evaluate an optional, more complex structure \textit{CAE with Generator}, which feeds the particle state to an additional generator network and increases state sensitivity. In this case, generator and decoder outputs are concatenated and fused with a final block of convolutions.
	During training (green), the target is a primed image with the same background as the input image but a different state.
	When deploying as part of a particle filter (purple), the target image is the same as the input, with the network conditioned on the queried particle state.
% 	In case of the variant \textit{CAE with Generator}, the output module also received inputs from the generator module that directly transforms the condition vector to an image shaped tensor.
	}
	\label{fig:cae_structure}
\end{figure*}

Let $X$ be the model of a system. $x$ is the state the system is currently in. Such a model is a simplified abstraction of the real world in which only those parts of the world are described that are relevant for a particular task. The state vector may contain heterogeneous elements such as position vectors, orientation quaternions, or categorical variables.
We call the unmodeled part of the reality $Z$ and $z$ the corresponding unmodeled state.  For instance, in a perception problem for robotic manipulation, $x$ could represent the pose of the object that needs to be grasped by a robot. In the same scenario, $z$ would correspond to the state of the background and surrounding objects. A measurement $y$ usually does not only depend on the modeled system state $x$ but also on the unmodeled state $z$. We suppose that there exists a function $f: (x,z) \to y$ that uniquely maps $x$ and $z$ to a measurement $y$.\\
Suppose the system is in the state $(x', z')$ and the measurement $y'$ is observed.
We want to infer $p(y'|x_i)$ for a set of particles $i=1,\dots, n$ by computing the similarity $\mathcal{L}$ of the corresponding measurements $y_i$ to $y'$.
We assume that $p(y'|x_i) \propto 1 / \mathcal{L}(y_i, y')$, given $z_i=z'$.\\
Since the measurements $y_i$ do not only depend on the known states $x_i$ but also the unmodeled part of the system, they cannot trivially be rendered with the function $f$ since $z'$ is usually not known.\\
We propose to use a conditional encoder-decoder network to render a measurement $y_i$ given the state $x_i$ without explicit knowledge of the unmodeled state $z'$.
Fig.~\ref{fig:cae_structure} shows the structure of the proposed method.
An encoder network $\phi$ extracts a latent code from an input measurement $y'$. The latent code resembles the unmodeled state $z'$. Together with the modeled state vector $x_i$, the decoder $\psi$ generates the measurement $y_i$.\\
In this work, we call the encoder-decoder combination Conditional Autoencoder (CAE), which is slightly inaccurate nomenclature since the learned network transcodes the tuple $(y', x_i)$ to the target $y_i$. 
\subsection{Primed Data Training}
Training the CAE is challenging since the encoder network has to learn how to extract a latent code that only describes the unmodeled state $z$ while ignoring the $x$ dependencies.\\
CVAEs achieve this separation between the latent code and the condition variable by imposing structure on the latent code: the encoder network does not directly infer the latent code but the parameters of a normal distribution $\mu$ and $\sigma$. The latent code $z$ is then sampled from the distribution $\bm{z} \sim \mathcal{N}(\mu, \sigma)$. A Kullback-Leibler divergence cost term penalizes deviations of $\bm{z}$ from the standard normal distribution.\\
This approach is not suitable for our problem. A CVAE only allows for sampling from the distribution of all possible unmodeled states $\bm{z}$ but does not extract the specific instance $z'$ we need to reconstruct $y_i$ from $x_i$.\\
Instead, we achieve the separation by training the networks with primed synthetic data: a synthetic data generator implements the function $f(x, \textit{z})$. $\textit{z}$ is some parametrization of $Z$ but typically not identical to the one that the encoder-decoder network learns. We use $f$ to generate two measurements: $y_1 = f(x_1, \textit{z})$ and $y_2 = f(x_2, \textit{z})$.
Both measurements are based on different modeled states but share the same unmodeled state. During training, we provide $y_1$ as an input to the encoder and $x_2$ as a condition variable. The training loss is the similarity $\mathcal{L}$ between the decoder output $\tilde{y}_2$ and the synthetic measurement $y_2$.
\begin{align}
    %\tilde{y}_2 &= \psi(\phi(y_1), x_2)\\
    \mathcal{L}_{training} &= \mathcal{L}(\tilde{y}_2 = \psi(\phi(y_1), x_2), y_2)
    \label{eqn:training_loss}
\end{align}
This encourages the encoder network to ignore the influence of $x_1$ on $y_1$ and only focus on the regression of $\textit{z}$ since this information is necessary to perform well in the task of generating $\tilde{y}_2$ with the decoder.
\section{Implementation}
\label{sec:implementation}
\subsection{Network Architecture}
\label{sub:network_architecture}
Two different Conditional Autoencoder networks have been evaluated.
The basic \textit{CAE} has an encoder and a decoder network. The encoder network compresses the input image with 6 convolutional layers followed by 3 linear layers all with Relu activations to a 64 dimensional latent vector. The decoder network reconstructs a depth image from the latent vector and the state vector with 6 deconvolutional layers.
The \textit{CAE with Generator} network shares the basic building blocks with the \textit{CAE} except that it has an additional generator module consisting of 7 convolutional layers. The generator output is combined with the decoder output in the output conv module with 2 convolutional layers. 
%Fig.~\ref{fig:cae_structure} visualizes the dataflow between the different modules.
% \begin{figure}[htp]
% 	\centering
% 	\includegraphics[width=\columnwidth]{figures/CAE_network.png}
% 	\caption{The encoder network compresses the input image into a latent vector. The latent vector is augmented with the condition vector (i.e. the particle state of a base filter). The augmented latent is then decompressed by the decoder and fed to the output module. In case of the variant \textit{CAE with Generator}, the output module also received inputs from the generator module that directly transforms the condition vector to an image shaped tensor.
% 		\todo[inline]{Draw better figure of the structure}}
% 	\label{fig:cae_network_structure}
% \end{figure}
The rationale behind the encoder and decoder structure is described in subsection \ref{ss:dmu}. The generator network supports the decoder in delivering high-quality reconstructions by increasing the emphasis on the modeled state $x$ dependency. Conceptually it can be viewed as a part of the decoder implementation.\\
Convolutional layers extract local features from their inputs or reconstruct pixels from those features in the case of deconvolution. However, they lack a sense of global positioning information in an image. We found that this increases the difficulty of creating good reconstructions. To alleviate that problem, we use Coordinate Convolutional (CoordConv) Layers that take the horizontal and vertical pixel coordinates as additional inputs \cite{liuIntriguingFailingConvolutional2018}.\\
The decoder and encoder use kernel size 4 and stride 2; the generator and output networks only employ kernel size 1 convolutions with stride 1.
\subsection{Training}
\label{sec:implementation_training}
A data loader has been implemented that generates training data in parallel to the training process. It uses Nvidia's Issac Sim\footnote{\url{https://developer.nvidia.com/isaac-sim}} to generate batches of depth image pairs $(y_1, y_2)$ with a resolution of $128\times256 ~ \texttt{pixel}$ along with their corresponding states.
The use of depth images over RGB data is motivated by recent works showing that depth images can transfer from simulation to real applications~\cite{danielczuk2019segmenting}, \cite{buch2017prediction}, \cite{wan2017crossing}.
Additionally, we generate segmentation masks $\upsilon_2$ for the image $y_2$ with problem-specific regions of interest labeled.\\
$1050$ samples are generated for each epoch which are reused $5$ times in shuffled order. The minibatch size is set to $35$.\\
As a loss function $\mathcal{L}$ we use the L1-norm between $y_1$ and $y_2$.
During experimentation, we noticed an imbalance in the number of pixels affected by the difference between $x_1$ and $x_2$ and the parts of the image that only depend on $z$ and are therefore equal between $y_1$ and $y_2$. To counteract the imbalance, we use the segmentation mask $\upsilon_2$ to compute the losses for labeled and unlabeled pixels separately. The total loss is the sum of both labeled and unlabeled loss with mean normalization applied over all pixels in a mini-batch for the separate loss terms. At test time, the segmentation masks are not available, and we compute the loss as described in \eqref{eqn:training_loss}.
The networks are trained on an Nvidia RTX 2080 Super GPU with an Adam optimizer \cite{ADAM2015} and a learning rate of $10^{-4}$ until convergence.
\begin{figure*}[htp]
	\centering
	\begin{subfigure}[t]{0.245\textwidth}
		\includegraphics[width=1.0\textwidth]{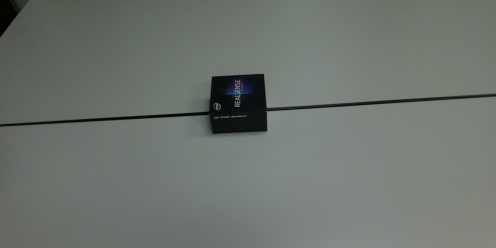}
		\caption{Scene}
		\label{fig:exp_a_scene}
	\end{subfigure}
	\hfill
	\begin{subfigure}[t]{0.245\textwidth}
		\includegraphics[width=1.0\textwidth]{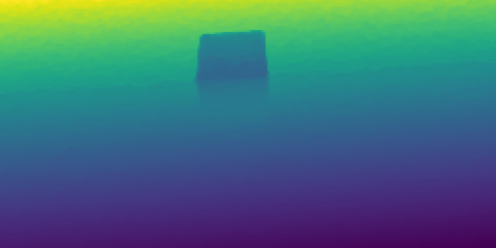}
		\caption{Input Depth Image}
		\label{fig:exp_a_input}
	\end{subfigure}
    \hfill
    \begin{subfigure}[t]{0.245\textwidth}
		\includegraphics[width=1.0\textwidth]{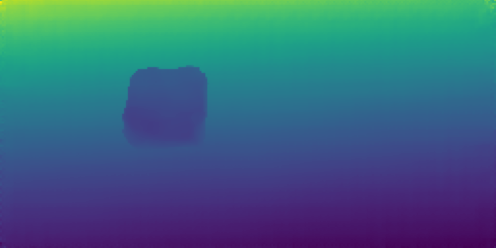}
		\caption{Reconstruction with $T_{gt} + \SI{10}{\centi\meter} \cdot (\vec{n}_{x, obj} + \vec{n}_{y, obj})$, \textit{CAE with Generator} model.}
		\label{fig:exp_a_reconstruction}
	\end{subfigure}
	\hfill	
	\begin{subfigure}[t]{0.245\textwidth}
		\includegraphics[width=1.0\textwidth]{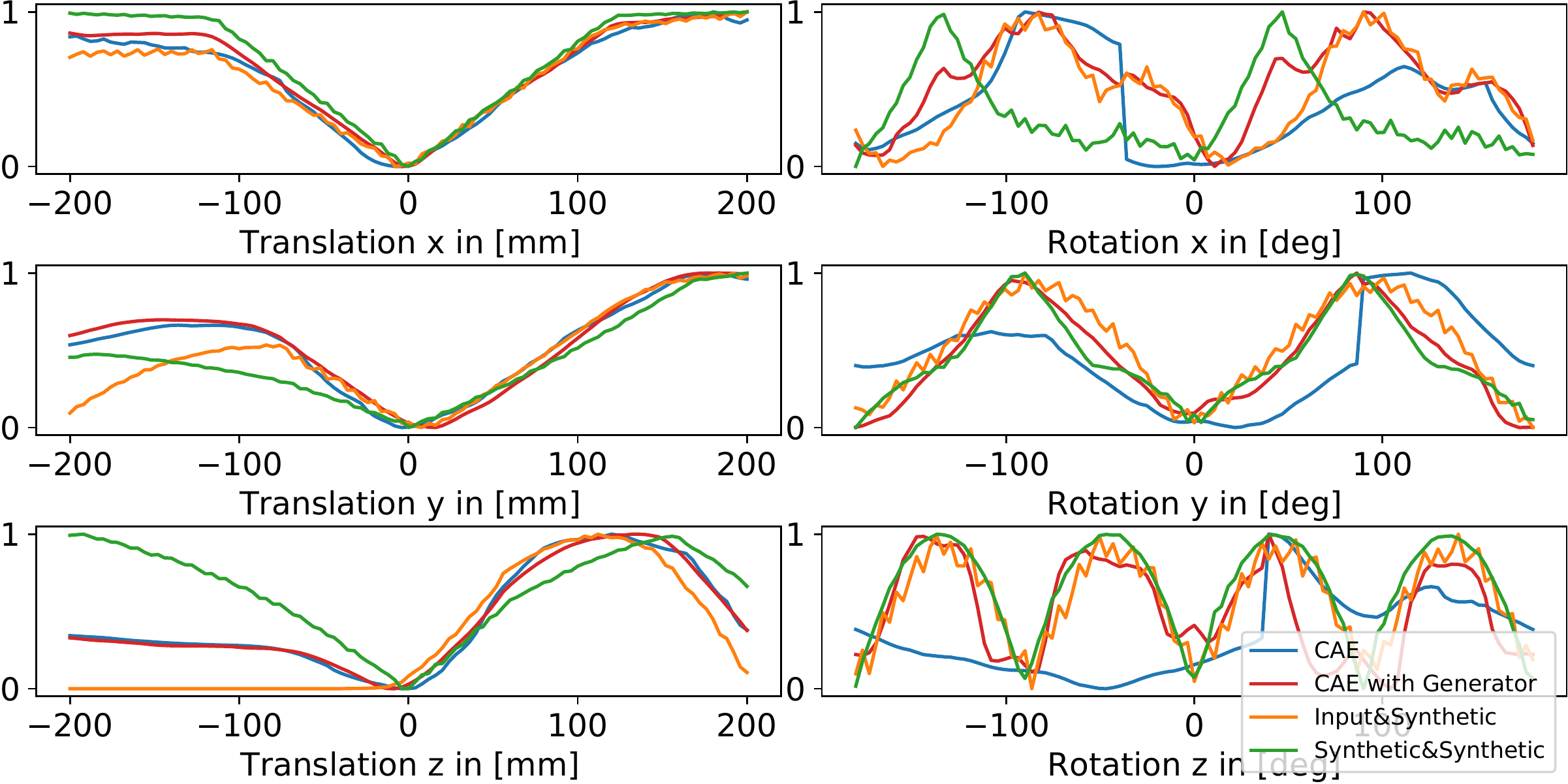}
		\caption{Likelihood Function}
		\label{fig:exp_a_conditional_likelihood_cae}
	\end{subfigure}
	\hfill
	\begin{subfigure}[t]{0.245\textwidth}
		\includegraphics[width=1.0\textwidth]{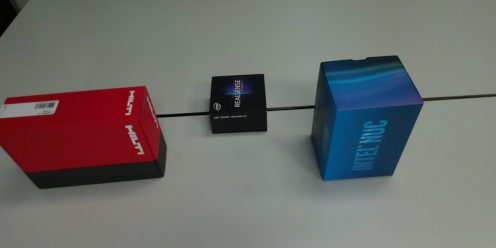}
		\caption{Scene}
		\label{fig:exp_b_scene}
	\end{subfigure}
	\hfill
	\begin{subfigure}[t]{0.245\textwidth}
		\includegraphics[width=1.0\textwidth]{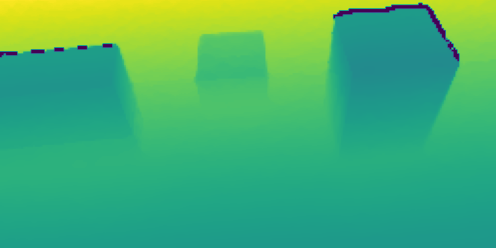}
		\caption{Input Depth Image. The black dots are missing sensor readings.}
		\label{fig:exp_b_input}
	\end{subfigure}
	\hfill
	\begin{subfigure}[t]{0.245\textwidth}
		\includegraphics[width=1.0\textwidth]{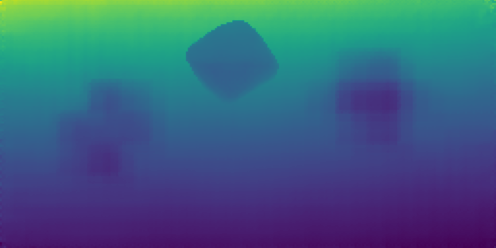}
		\caption{Reconstruction with $T_{gt}$ rotated by $\ang{45}$ around $\vec{n}_{z, obj}$, \textit{CAE with Generator} model.}
		\label{fig:exp_b_reconstruction}
	\end{subfigure}
	\hfill
	\begin{subfigure}[t]{0.245\textwidth}
		\includegraphics[width=1.0\textwidth]{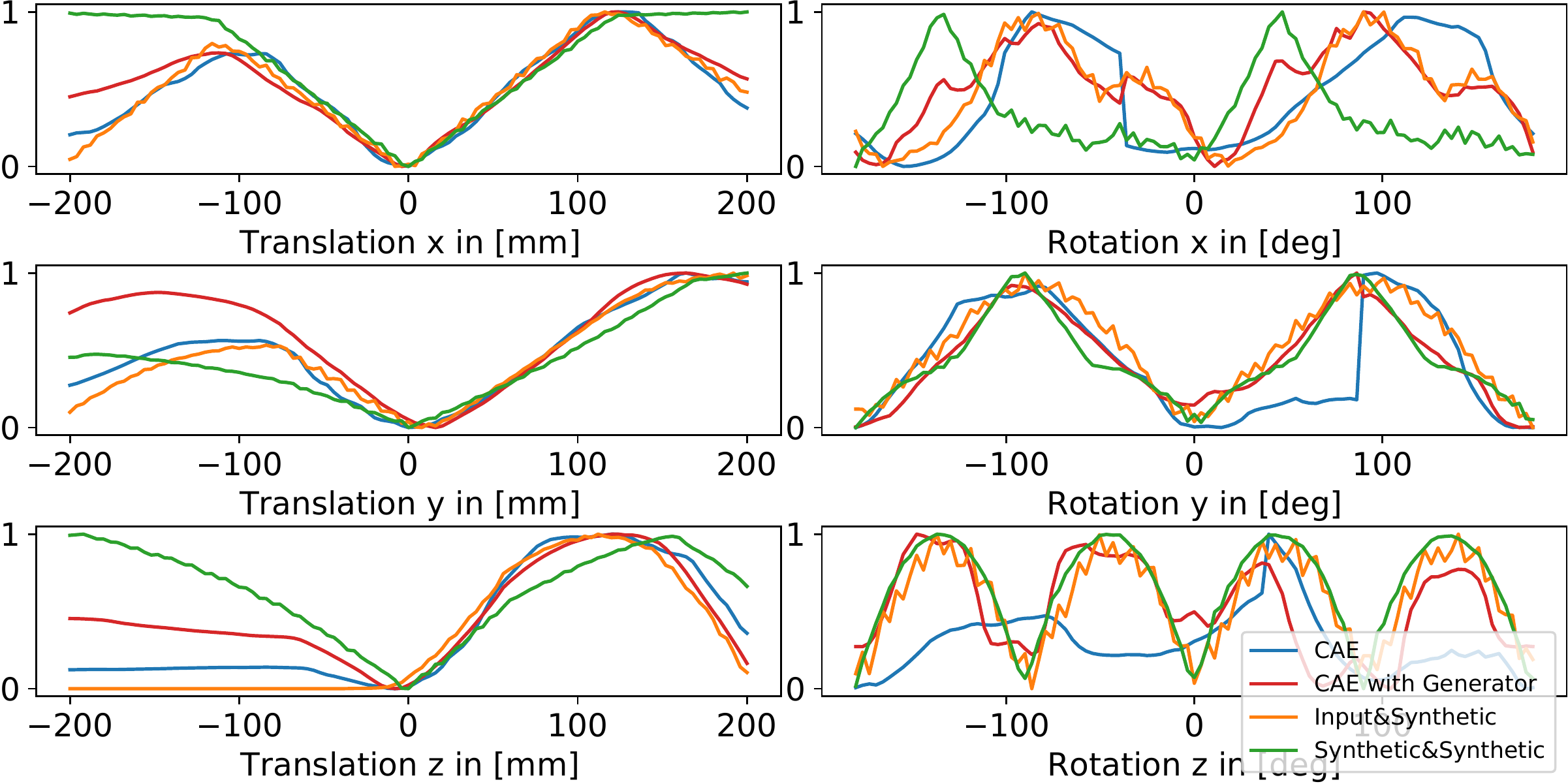}
		\caption{Likelihood Function}
		\label{fig:exp_b_conditional_likelihood_cae}
	\end{subfigure}
	\hfill
	\begin{subfigure}[t]{0.245\textwidth}
		\includegraphics[width=1.0\textwidth]{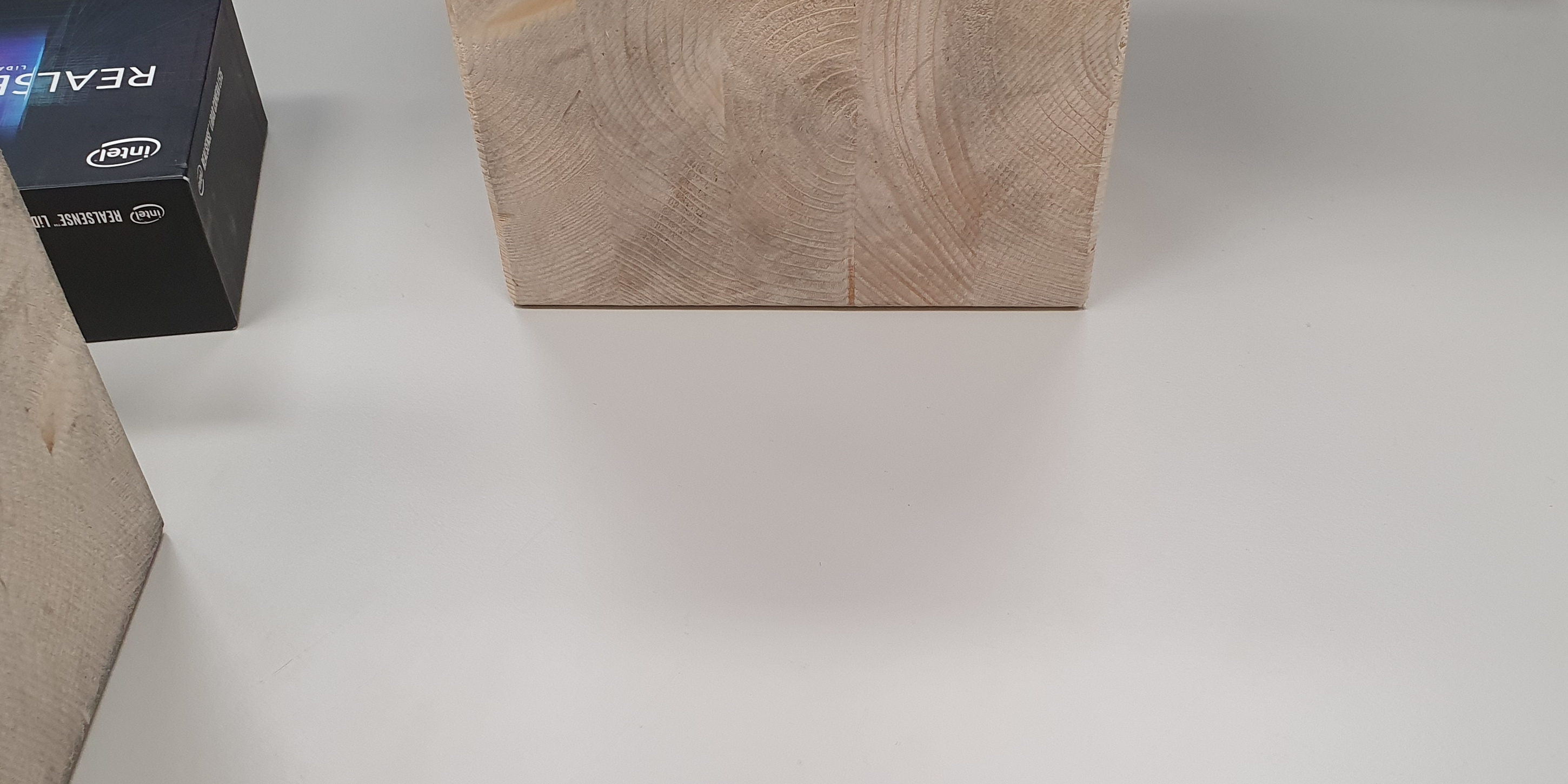}
		\caption{Scene}
		\label{fig:exp_c_scene}
	\end{subfigure}
	\hfill
	\begin{subfigure}[t]{0.245\textwidth}
		\includegraphics[width=1.0\textwidth]{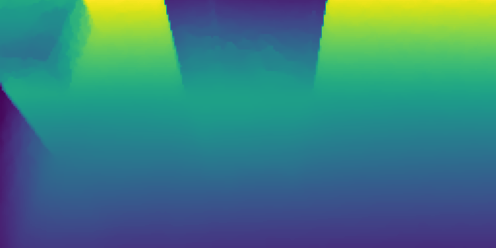}
		\caption{Input Depth Image}
		\label{fig:exp_c_input}
	\end{subfigure}
	\hfill
	\begin{subfigure}[t]{0.245\textwidth}
		\includegraphics[width=1.0\textwidth]{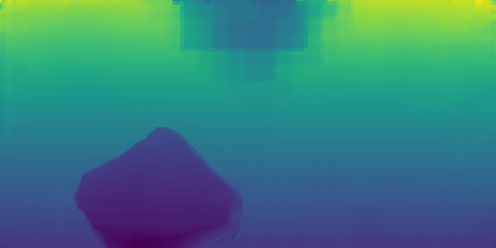}
		\caption{Reconstruction with $T_{gt}$ translated by $\SI{20}{\centi\meter} \cdot (-\vec{n}_{x, obj} + \vec{n}_{y, obj})$ and rotated by $\ang{45}$ around $\vec{n}_{z, obj}$, \textit{CAE with Generator} model.}
		\label{fig:exp_c_reconstruction}
	\end{subfigure}
	\hfill
 	\begin{subfigure}[t]{0.245\textwidth}
 		\includegraphics[width=1.0\textwidth]{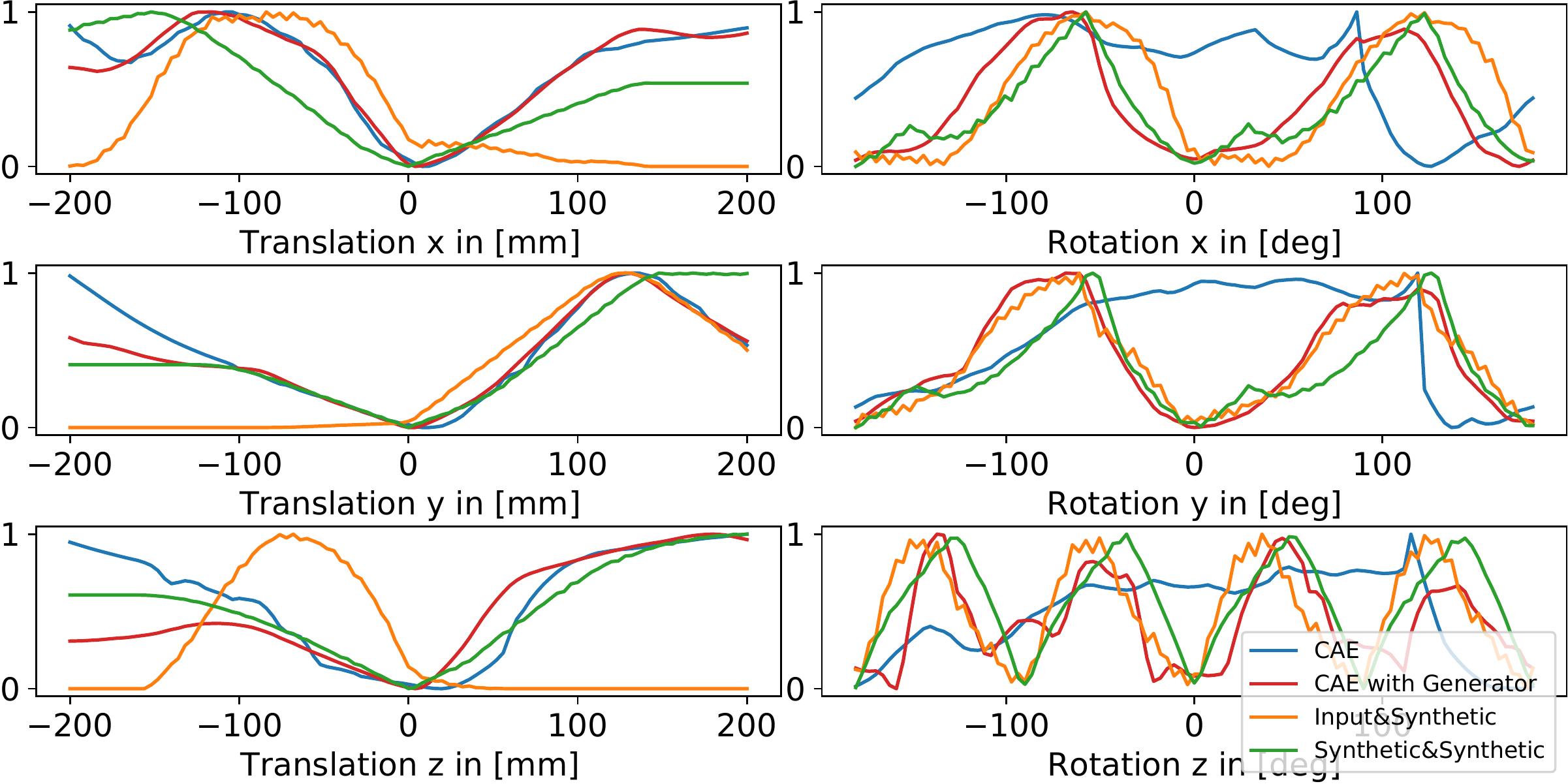}
 		\caption{Likelihood Function}
 		\label{fig:exp_c_conditional_likelihood_cae}
 	\end{subfigure}
	\begin{subfigure}[t]{0.245\textwidth}
		\includegraphics[width=1.0\textwidth]{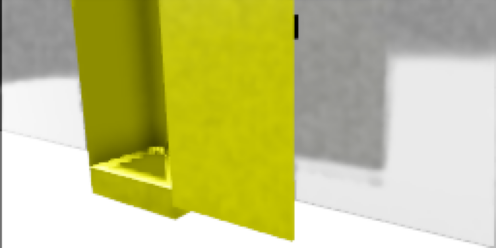}
		\caption{Synthetic RGB Scene Image}
		\label{fig:cabinet_exp_synthetic_scene}
	\end{subfigure}
	\hfill
	\begin{subfigure}[t]{0.245\textwidth}
		\includegraphics[width=1.0\textwidth]{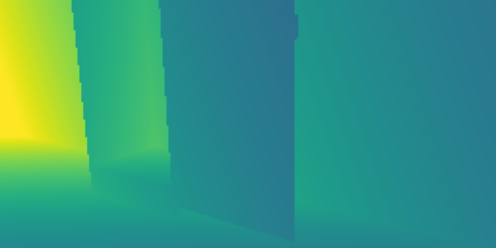}
		\caption{Synthetic Input Depth Image}
		\label{fig:cabinet_exp_synthetic_input_image}
	\end{subfigure}
	\hfill
	\begin{subfigure}[t]{0.245\textwidth}
		\includegraphics[width=1.0\textwidth]{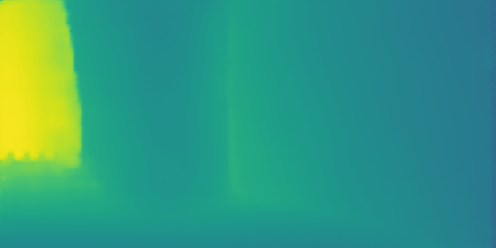}
		\caption{Reconstruction with $\alpha_{gt} - 180 \ \textrm{deg}$}
		\label{fig:cabinet_exp_synthetic_reconstruction}
	\end{subfigure}
	\hfill
	\begin{subfigure}[t]{0.245\textwidth}
		\includegraphics[width=1.0\textwidth]{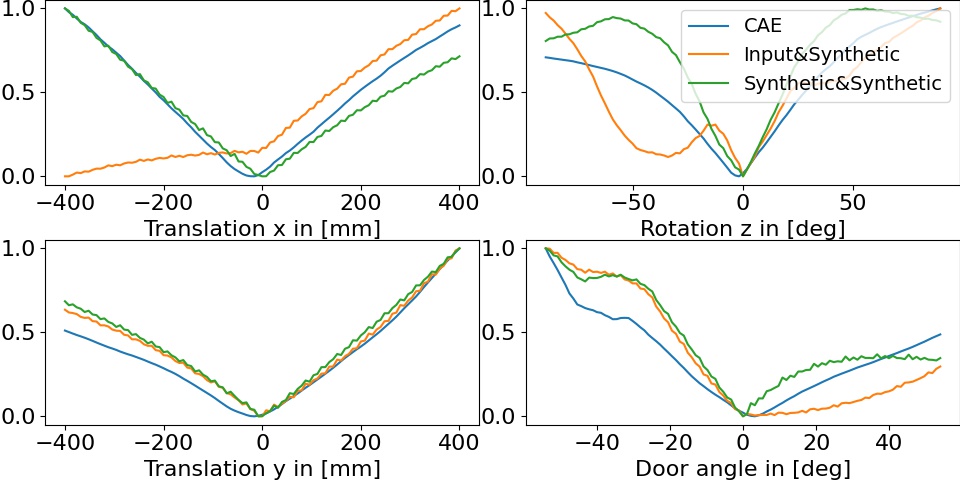}
		\caption{Likelihood Function}
		\label{fig:cabinet_exp_synthetic_conditional_likelihoods}
	\end{subfigure}
	\hfill
	\begin{subfigure}[t]{0.245\textwidth}
		\includegraphics[width=1.0\textwidth]{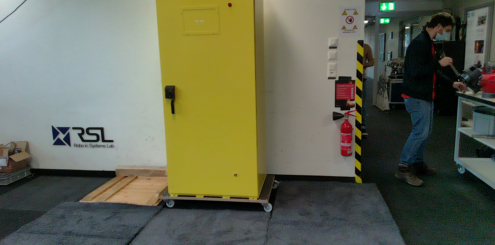}
		\caption{Scene}
		\label{fig:cabinet_exp_real_scene}
	\end{subfigure}
	\hfill
	\begin{subfigure}[t]{0.245\textwidth}
		\includegraphics[width=1.0\textwidth]{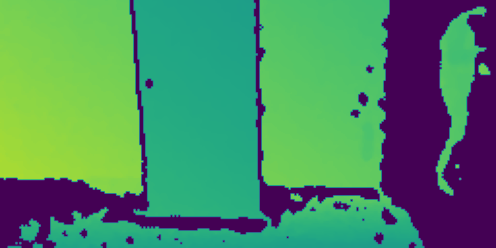}
		\caption{Input Depth Image}
		\label{fig:cabinet_exp_real_input_image}
	\end{subfigure}
	\hfill
	\begin{subfigure}[t]{0.245\textwidth}
		\includegraphics[width=1.0\textwidth]{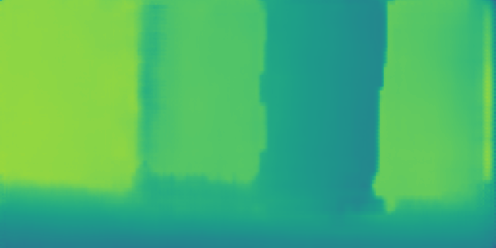}
		\caption{Reconstruction with $\alpha_{gt} + 140 \ \textrm{deg}$}
		\label{fig:cabinet_exp_real_reconstruction}
	\end{subfigure}
	\hfill
	\begin{subfigure}[t]{0.245\textwidth}
		\includegraphics[width=1.0\textwidth]{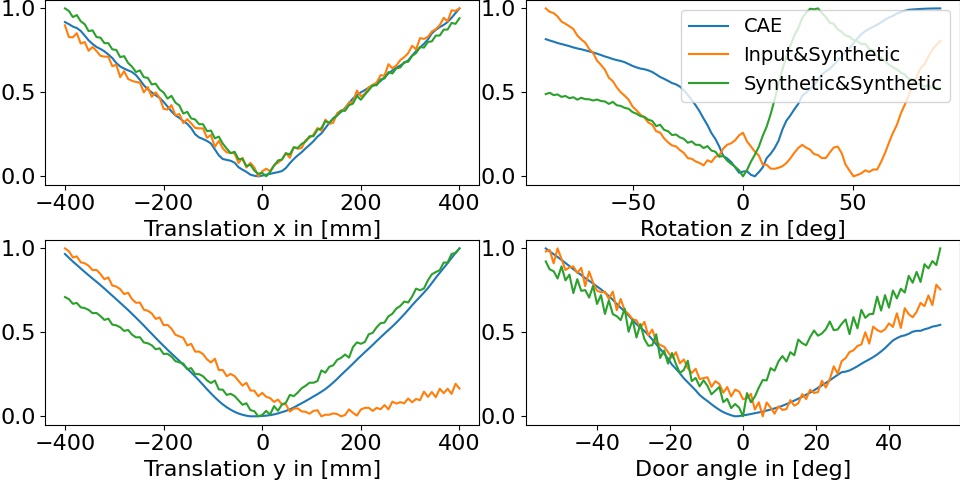}
		\caption{Likelihood Function}
		\label{fig:cabinet_exp_real_conditional_likelihoods}
	\end{subfigure}
	\caption{
	Top 3 rows: Symmetric Object Experiment results. Bottom 2 rows: Results of the switchboard cabinet experiment.\\
	Column 1 gives an overview of the scene. Column 2 shows the recorded depth images. In column 3, reconstructions of the depth images with state vectors deviating from the ground truth are shown. Column 4 shows the loss $\mathcal{L}$ evaluated as a function of the state for the trained models \textit{CAE} (blue) and \textit{CAE with Generator} (red) and the two benchmark quantities \textit{Input\&Synthetic} (orange) and \textit{Synthetic\&Synthetic} (green).
	For the symmetric pose experiments the state vector $x$ is the object pose in camera frame.
	In the switchboard cabinet experiment, the state vector is the translation of the cabinet, the yaw angle, and the door opening angle.}
	\label{fig:likelihood_plots}
\end{figure*}
\section{EXPERIMENTS}
\label{sec:experiments}
We present several experimental evaluations of the proposed method. In Sec.~\ref{subsec:object_pose_measurement_update}, we demonstrate how the method can be used to infer the observation probability of a measurement when the system describes the pose of an object.
Even though object pose estimation is not the main focus of this work and as a depth-only method it cannot compete with modern RGB-D detectors, we included this experiment to show that DMU's likelihood estimation can handle ambiguities from symmetries and occlusions nicely.
In Sec.~\ref{subsec:particle_filter_integration}, we describe the integration of DMU into a particle filter to estimate the pose of an object in a scene with occluding obstacles.
In Sec.~\ref{subsec:benchmark_comparison}, DMU's ability to infer the articulation state of an object is compared to A-SDF on the RBO dataset.
Sec.~\ref{subsec:articulated_object_measurement_update} demonstrates the capabilities of DMU in a more complex perception problem, where both the pose and door opening angle of a switchboard cabinet are used as condition states for the CAE network.
In all the experimental scenarios we evaluate on real-world depth images.% recorded with an Intel Realsense L515 solid state LIDAR.
%We show how models that are only trained on synthetic data transfer to real-world data.
\subsubsection*{Likelihood Function Evaluation}
Since the learned CPDF $p(y|x)$ is high dimensional, it is difficult to visualize as a whole.
Instead, we evaluate the learned models as likelihood functions. We treat the state vector $x$ as parameters for a given input image and evaluate the likelihood function along the coordinate axes around the ground truth state.
The learned likelihoods are compared to two benchmark quantities that we call \textit{Input\&Synthetic} and \textit{Synthetic\&Synthetic}:\\
For the first benchmark, \textit{Input\&Synthetic}, a synthetic image $y_{syn}(x)$ with the query state $x$ is rendered. The image does not have a background; all non-foreground pixels are set to the maximum depth value.
We compute the pixel-wise minimum between the input image and the synthetic image to copy the background and possible occlusions of the input image into the synthetic image.
This combined image is used as a replacement for the learned reconstruction image $\tilde{y}$ in the loss computation:
\begin{equation}
    \textit{Input\&Synthetic}(y, x) = \mathcal{L}(y, min(y, y_{syn}(x)))
\end{equation}
The benchmark generally performs better than a simpler version in which the loss between the input image $y$ and the synthetic image $y_{syn}(x)$ is computed, since the simple version has no notion of occlusions.\\
The second benchmark, \textit{Synthetic\&Synthetic}, compares a rendering of the ground truth state $x_{gt}$ with the rendered scene at query state $x$:
\begin{equation}
    \textit{Synthetic\&Synthetic}(x_{gt}, x) = \mathcal{L}(y_{syn}(x), y_{syn}(x_{gt}))
\end{equation}
%While the second benchmark is the best possible result that the learned model could achieve under ideal conditions with no occlusions, it cannot replace the learned model in a real-world deployment since it assumes knowledge of the ground truth state.\\
The second benchmark is the best possible result that the learned model could achieve under ideal conditions with no occlusions. It requires knowledge of the ground truth state $x_{gt}$, which we determined for the evaluation experiments.
The first benchmark does not require this ground truth knowledge and can therefore compete with the learned model in application scenarios.
\subsection{Symmetric Object With Occlusions}
\label{subsec:object_pose_measurement_update}
For this experiment we examine the system $X_{object}$ in which $x$ describes the pose of a box in camera frame with dimensions $\SI{11}{\centi\meter}\times\SI{11}{\centi\meter}\times\SI{6}{\centi\meter}$. The rotation is encoded with a quaternion to avoid discontinuities in the state parameterization.
The box shape was chosen to demonstrate the strength of our method in handling symmetric objects naturally.
Inferring the pose of a symmetric object represents a particularly challenging problem, as documented in literature \cite{pitteriObjectSymmetries6D2019}, \cite{corona2018pose}, \cite{richter2020handling}. This is because the likelihood to be estimated is multi-modal, something that direct methods for object pose inference do not explicitly handle. In fact, such methods either infer one possible rotation arbitrarily \cite{xiang2017posecnn} or require symmetry labeled training data \cite{pitteriObjectSymmetries6D2019}.
Objects of other shapes can easily be considered by including their meshes into the synthetic data loader.
In addition to the target object, a plane and up to three boxes are added to the scene. They constitute the unmodeled part of the system. The poses of the plane and boxes and the boxes sizes are randomized. All additional objects may occlude the target object partially or entirely.
\subsubsection*{Results}
Fig.~\ref{fig:exp_a_scene}-\ref{fig:exp_c_conditional_likelihood_cae} shows the results of the proposed experiments.
The trained models are evaluated on several real-world depth images. In those images, the box is placed on a table. We present the results of scenes with increasing difficulty. In the first scene, the object is placed at the center of the scene, and no obstacles are present. In the second scene, two obstacles are placed next to the target object. The obstacles are also box-shaped but have different sizes than the target object. In the third scene, the target object is only partially visible, and there are two obstacles present.
In Fig.~\ref{fig:exp_a_reconstruction},\ref{fig:exp_b_reconstruction},\ref{fig:exp_c_reconstruction}, we present examples of reconstructed input depth images from Fig.~\ref{fig:exp_a_input},\ref{fig:exp_b_input},\ref{fig:exp_c_input} with different condition vectors.
% In the presented cases, the cube can be reconstructed at the desired position by both models.
% %The objects vanish at their original pose from the input images.
% The networks are able to extrapolate the background to complete the depth images at the original cubes' locations.
% The reconstruction with the basic CAE model in figure \ref{fig:exp_a_reconstruction} is considerably more blurry than the reconstructions with the \textit{CAE with Generator} model in figures \ref{fig:exp_b_reconstruction} and \ref{fig:exp_c_reconstruction}.
\begin{figure*}[htp]
	\centering
	\begin{subfigure}[t]{0.24\textwidth}
		\includegraphics[width=1.0\textwidth]{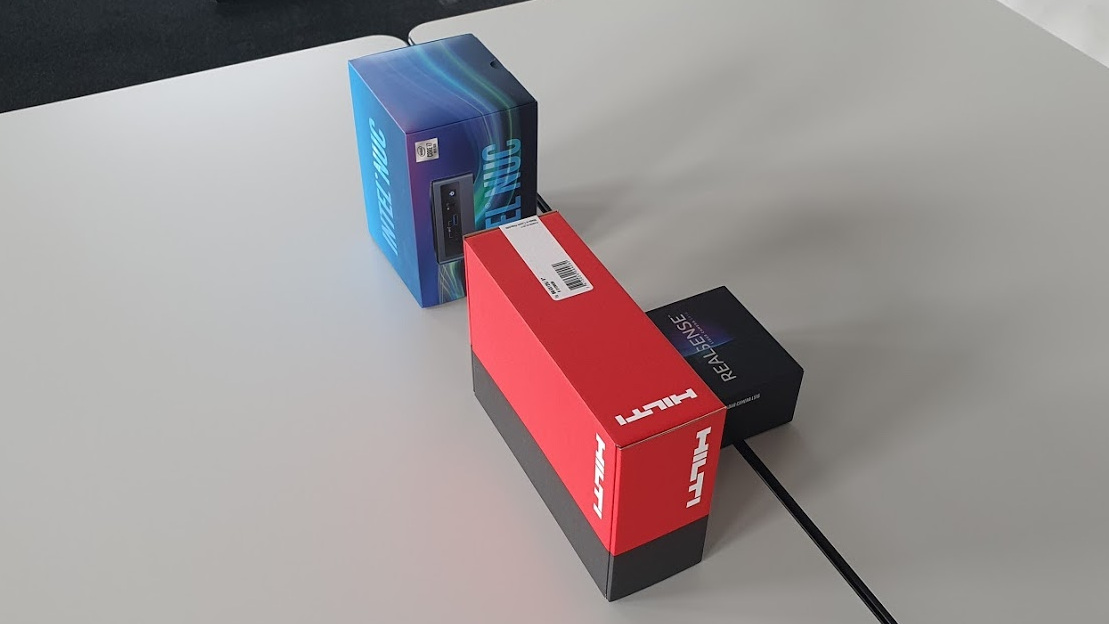}
		\caption{Scene overview, initial state.}
		\label{fig:pfscene}
	\end{subfigure}
	\hfill
	\begin{subfigure}[t]{0.24\textwidth}
		\includegraphics[width=1.0\textwidth]{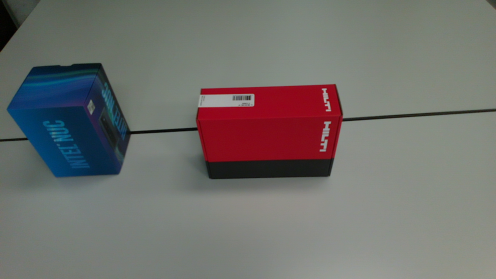}
		\caption{Two obstacles are present}
		\label{fig:pf2_color}
	\end{subfigure}
	\hfill
	\begin{subfigure}[t]{0.24\textwidth}
		\includegraphics[width=1.0\textwidth]{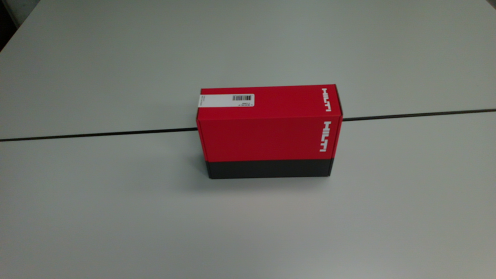}
		\caption{One obstacle has been removed.}
		\label{fig:pf1_color}
	\end{subfigure}
	\hfill
	\begin{subfigure}[t]{0.24\textwidth}
		\includegraphics[width=1.0\textwidth]{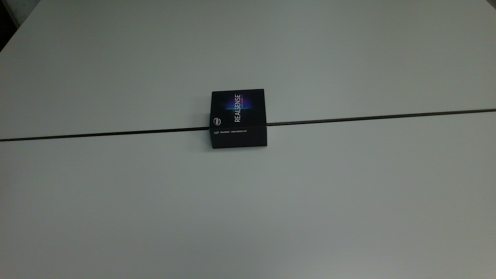}
		\caption{The target object is visible now.}
		\label{fig:pf0_color}
	\end{subfigure}\\
	\begin{subfigure}[t]{0.24\textwidth}
		\includegraphics[width=1.0\textwidth]{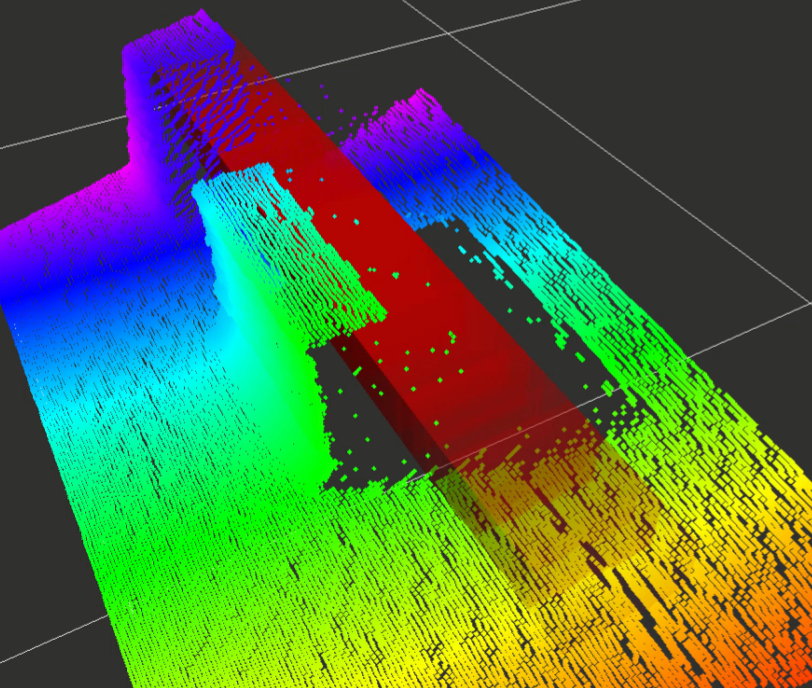}
		\caption{Initial condition, no DMU has been applied yet.}
		\label{fig:pf_initial}
	\end{subfigure}
	\hfill
	\begin{subfigure}[t]{0.24\textwidth}
		\includegraphics[width=1.0\textwidth]{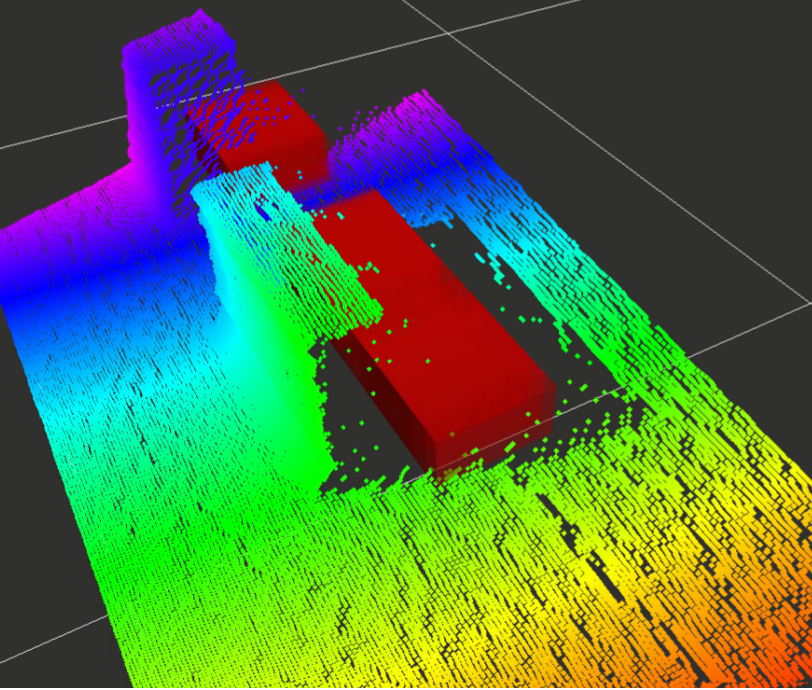}
		\caption{The particle distribution becomes bimodal. The filter assumes that the box is hidden behind one of the obstacles.}
		\label{fig:pf2}
	\end{subfigure}
	\hfill
	\begin{subfigure}[t]{0.24\textwidth}
		\includegraphics[width=1.0\textwidth]{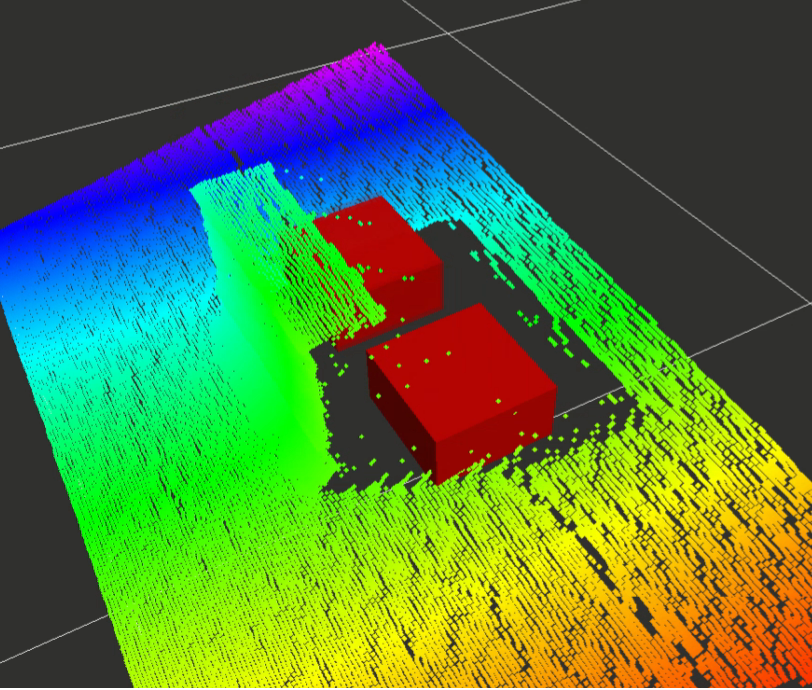}
		\caption{After removing one obstacle, the particles behind that obstacle annihilate.}
		\label{fig:pf1}
	\end{subfigure}
	\hfill
	\begin{subfigure}[t]{0.24\textwidth}
		\includegraphics[width=1.0\textwidth]{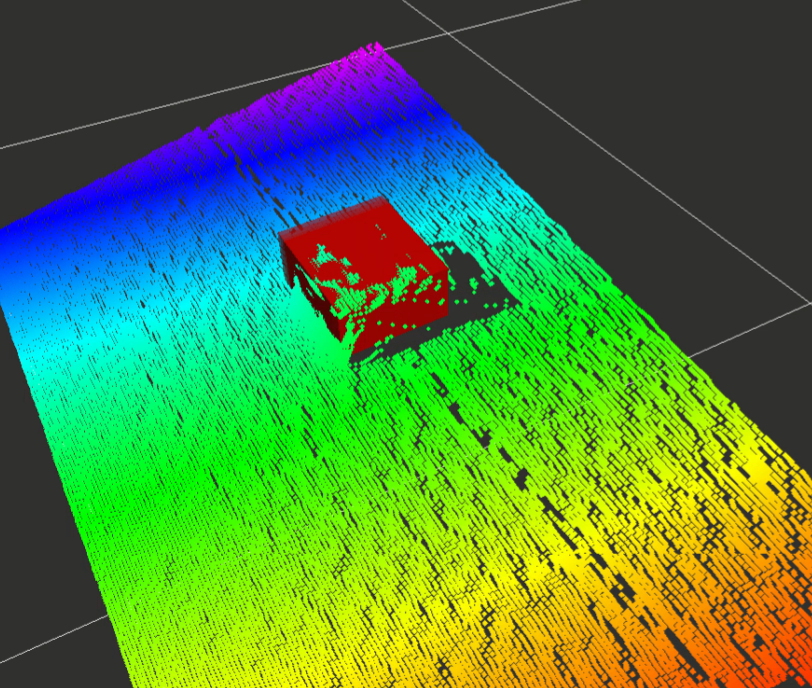}
		\caption{The target object is visible. The filter converges to the ground truth state.}
		\label{fig:pf0}
	\end{subfigure}
	\caption{Selected stills from a particle filter implementation with deep measurement updates.
		A box to localize is initially hidden behind one of two obstacles. Fig.~\ref{fig:pfscene} shows an overview of the scene, Fig.~\ref{fig:pf2_color},\ref{fig:pf1_color},\ref{fig:pf0_color} show the changing scene from the viewpoint of the sensor.
		The point clouds are 3D visualizations of the recorded depth images. The current particle states are visualized as red semi-transparent boxes.
		As long as the target object is not visible, DMU assigns higher probabilities to the particles hidden behind the obstacles. While obstacles are consequently removed, the state estimate is refined until the filter converges to the ground truth state when the object becomes visible.}
	\label{fig:pf}
\end{figure*}
In the presented cases, the cubes can be reconstructed at the desired poses.
The networks are able to extrapolate the background to complete the depth images at the original cubes' locations.
We noticed that the reconstructions with the basic \textit{CAE} model were considerably more blurry than the presented results of the \textit{CAE with Generator} in Fig.~\ref{fig:exp_a_reconstruction},\ref{fig:exp_b_reconstruction},\ref{fig:exp_c_reconstruction}.
This suggests that the additional component in the network architecture makes the trained model more expressive.
Both models are able to reconstruct the background plane from the input images well. The obstacles in \ref{fig:exp_b_reconstruction} and \ref{fig:exp_c_reconstruction} are also visible in the generated images, but they are less sharp than the target objects. During the development process of the method, we saw that the background reconstruction quality was higher when training with a simple reconstruction loss instead of using the weighted loss function on labeled pixels, described in Sec.~\ref{sec:implementation_training}. However, for a good likelihood estimation the accurate reconstruction of the target object is of greater importance, so we compromised in the background reconstruction abilities.
Imperfections in the reconstruction of the background can lead to a constant bias on the $\mathcal{L}_1$ loss for each queried state vector. Scaling the loss values to unit range can alleviate the bias.
As shown in Figs.~\ref{fig:exp_a_conditional_likelihood_cae},\ref{fig:exp_b_conditional_likelihood_cae},\ref{fig:exp_c_conditional_likelihood_cae}, the likelihood functions of the learned models have their minima at the ground truth object states for the translational directions.
In all cases, these minima are global among the queried states. 
%which suggests that a Bayesian filter would converge to the ground truth state.
The asymmetric shape of the loss functions correlates to the change of scale when the object is moved along the axes.
The same effect can be observed in the \textit{Synthetic\&Synthetic} benchmark.
In general, the shape of the loss along the axes follows the \textit{Synthetic\&Synthetic} benchmark well, suggesting that the model performs well on the posed problem.\\
Compared to the benchmark \textit{Input\&Synthetic}, it stands out that our method does not suffer from self occlusions while the benchmark does. Especially for translations along the object's z-axis, the input image fully occludes the synthetic image in the benchmark case while the learned model can remove the box at its original pose.\\
The loss function exhibits symmetries when evaluating it along the rotational directions for the test object. Since the height (z-direction) differs from its depth (x-direction) and width (y-direction), the depth images should be identical for rotation of $\SI{180}{\degree}$ around x and y-axis and of $\SI{90}{\degree}$  around z-axis.
The expected symmetries can be observed with the \textit{CAE with Generator} model for rotations in directions.
The basic \textit{CAE} model cannot produce a reasonable likelihood function for rotations around the defined axes.
We hypothesize that the reconstruction of objects with the correct orientation is harder to achieve than with the correct translation since a smaller change in loss is observed when changing orientation compared to translation.
%\todo[inline]{Report scaling factors for plots}
%During the development of the final architecture of the CAE we noticed this effect and added the generator module and the normalization scheme with segmentation masks. Without those additions the object's orientation was indistinguishable while the translation problem could be solved with the simpler model.
\subsection{Particle Filter Integration}
\label{subsec:particle_filter_integration}
We integrate the DMU model with a particle filter implementation.
%The particle filter uses the Raisim simulator \cite{raisim} for physically accurate process updates.
DMU is used to compute weights for the re-sampling step during the measurement update. The normalizing factor $\eta$ is computed as the sum of all particle weights.
%This constraints the simulated box poses to the surface of the table.
%While the scene is static, a 6-DOF manipulator moves the camera across the scene. At the beginning of each recording the target object is fully occluded and it is only visible to the camera at a later stage. The provided supplementary video \footnote{\url{link.to.video}} features the recorded datasets.
A static depth camera observes a scene with two obstacles and a target object placed on a table. At the beginning of the experiment, the target object is hidden behind one of the obstacles. The two obstacles are then removed one by one such that the target object becomes visible.
\subsubsection*{Results}
The accompanying video\footnote{\url{https://youtu.be/MtTNIRcKBbk}} features the described experiment.
Fig.~\ref{fig:pf} shows selected stills from one experiment. For better visibility, we chose a one-dimensional initial distribution with 100 particles, while for other experiments featured in the video, the particles are randomly placed on the table.
After a few DMU steps, the particle distribution becomes bimodal with clusters of particles behind the two obstacles. Even though the target box is not visible yet, DMU can leverage the knowledge of observed free space to eliminate some particles.
After removing the first obstacle and revealing that the target has not been hidden here, the particle distribution collapses, and only the particles behind the second obstacle remain.
Finally, the target cube is visible in the depth image, and the filter converges to the ground truth box location.
\subsection{Articulation State Estimation}
\label{subsec:benchmark_comparison}
We compare DMU's ability to infer the articulation state of an object to A-SDF as a baseline solution. The comparison is made on the \textit{laptop} sequences of the RBO dataset. For each sequence, 10 depth images were selected by the authors of A-SDF for evaluation.
We train the DMU network on synthetic depth images with a laptop placed on a table with varying positions, orientations, and opening angles. The system state $x$ is the opening angle, and the position and orientation are part of the unmodeled state.
For each image of the test sequences, we compute the likelihood of $100$ possible joint angles. The angle with the highest likelihood is the estimate.
\subsubsection*{Results}
\begin{figure}[htp]
    \begin{subfigure}{1.0\columnwidth}
        \includegraphics[width=1.0\linewidth]{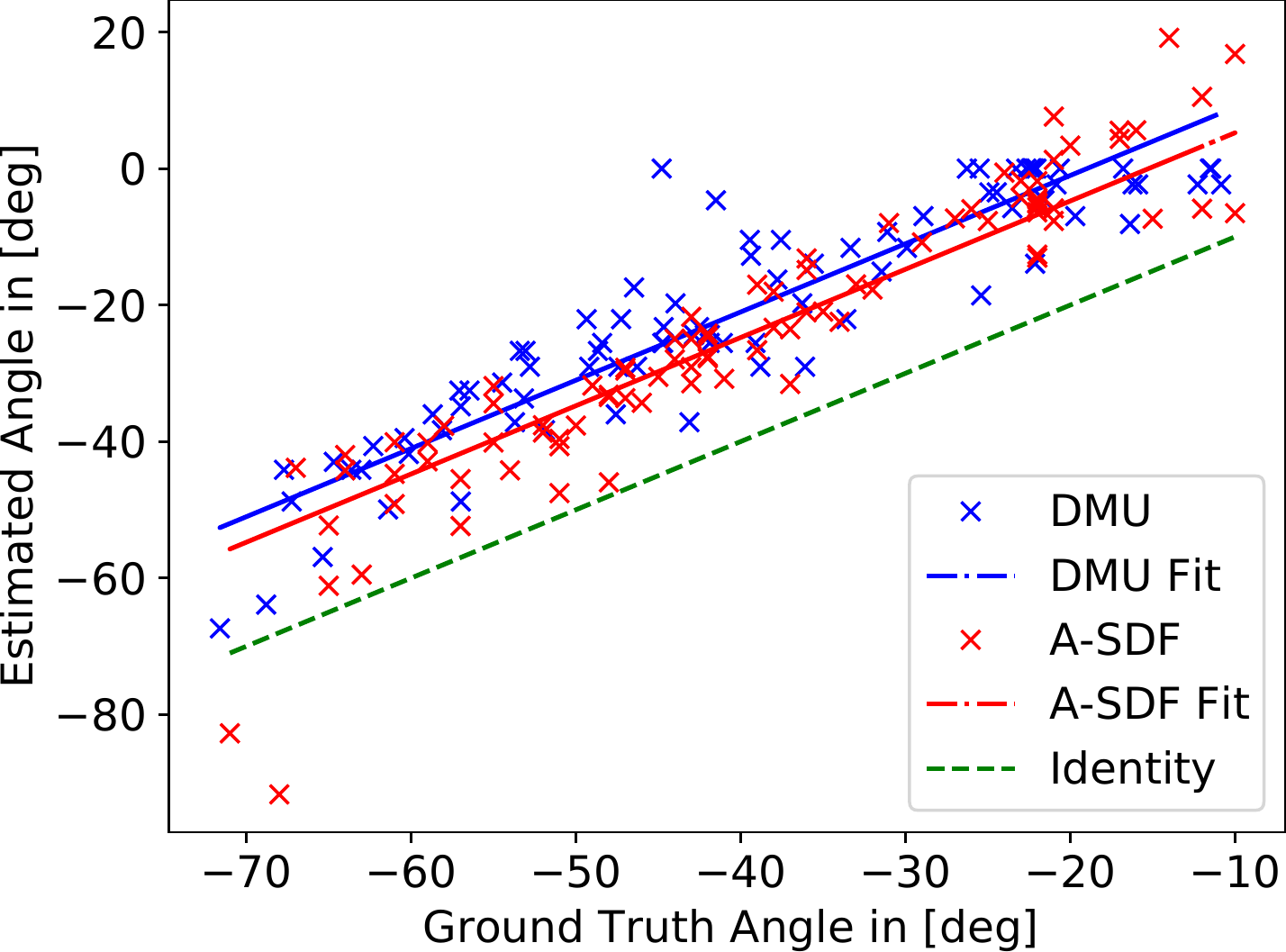}
        \caption{Scatter plot of estimated articulation angles over ground truth angles. Bias-compensated fits are visualized for both DMU and A-SDF.}
        \label{fig:rbo_scatter}
    \end{subfigure}\\
    %\hfill
    \begin{subfigure}{1.0\linewidth}
        \centering
        \sisetup{group-four-digits}
        \begin{tabular}{l c c }
            \toprule
                & DMU       & A-SDF       \\
            \midrule
        MAE & $\SI{20.23}{\degree}$     & -     \\
        MAE A-SDF subset & $\SI{26.63}{\degree}$     & $\SI{17.20}{\degree}$   \\
        MAE visualized subset  & $\SI{18.99}{\degree}$   & $\SI{15.91}{\degree}$     \\
        MAE unbiased  & $\SI{5.11}{\degree}$   & $\SI{5.29}{\degree}$     \\
        Inference time & $\SI{25(1)}{\milli\second}$ & $\SI{33(0)}{\second}$   \\
            \bottomrule
        \end{tabular}
    \caption{Reported values: Mean average error (MAE) on the entire dataset (4142 images), a  subset of 190 images picked by the authors of A-SDF, the subset of 90 we have picked for visualization and the average inference time per depth image.}
    \label{tab:rbo}
    \end{subfigure}
\caption{Joint angle estimates on \textit{laptop} sequences of RBO dataset. }
\label{fig:rbo}
\end{figure}

Fig.~\ref{fig:rbo_scatter} shows the estimated articulation states computed with both DMU and A-SDF for the sequences ${1, 2, 3, 5, 9, 11, 12, 13, 19}$. Tab.~\ref{tab:rbo} reports the mean average error (MAE) on the entire dataset and different subsets.
While A-SDF is generally more accurate in articulation angle estimation, the inference time of our method is three orders of magnitude smaller, making it practical for real-time applications.
We have observed an estimation bias for both methods with respect to the ground truth labels. Visual inspection of the depth images from the dataset and synthetic data generated by the simulator lets us conclude that the ground truth labels were biased. We report the bias compensated MAE as well and see similar accuracies for DMU and A-SDF.
%In the failure cases of DMU, we often found that the object pose could not be extracted from the depth image correctly. We believe that providing segmentation masks as an additional input to the CAE might help to improve the detection rate.
%Compared to the ground truth labels, both methods perform similarly ($RMSE_{DMU}=\SI{20.15}{\degree}$;  $RMSE_{A-SDF}=\SI{17.08}{\degree}$). The fact that both methods overestimate the articulation angle by a similar offset suggests that there is a constant bias on the ground truth labels. Visual inspection of depth images generated by the simulator support this hypothesis. We therefore decided to also report bias-compensated errors ($RMSE_{DMU}=\SI{6.75}{\degree}$;  $RMSE_{A-SDF}=\SI{7.70}{\degree}$). On this metric, both methods perform equally well, too.
%Evaluated on all sequences, A-SDF outperforms DMU ($RMSE_{DMU}=\SI{36.16}{\degree}$;  $RMSE_{A-SDF}=\SI{17.20}{\degree}$). In the failure cases of DMU, it is unable to infer the laptop's pose in cluttered scenes correctly and hence also fails to estimate the correct articulation state. We believe that providing segmentation masks generated from RGB images with a detector like Mask R-CNN \cite{heMaskRCNN2018} would improve performance here.\\
%In terms of inference time, DMU clearly outperforms A-SDF. On a workstation computer with a RTX 2080 Super GPU, DMU takes $\SI{25(1)}{\milli\second}$ in contrast to $\SI{33(0)}{\second}$ for A-SDF to process a single image.
\subsection{Articulation State and Object Pose Estimation}
\label{subsec:articulated_object_measurement_update}
In this experiment, we evaluate the CAE model in a scenario where the network condition variables are the pose and the door opening angle of the switchboard cabinet shown in Fig~\ref{fig:cabinet_exp_real_scene}. 
% The capability of estimating both the pose and the internal state of an object is especially useful in robotics, where the object state is defined based on the characteristics of the task of interest. 
In particular, estimating the state of a door represents a common challenge in the current robotics manipulation research with increasing attention in recent years \cite{arduengo2019robust}, \cite{mittal2021articulated}.
For this experiment, the modeled state of the system is defined as $x = [p_x, p_y, \theta_z, \alpha]$, where $p_x, p_y$ are the planar position coordinates of the cabinet in a camera-fixed gravity-aligned coordinate frame, $\theta_z$ is the rotation of the cabinet around the vertical axis of that frame, and $\alpha$ is the door opening angle. Following the notation introduced in Sec~\ref{ss:dmu}, the unmodeled state of the system is represented by the walls and floor in the background image.
For training, the CAE model without the additional generator network has been used. During evaluation on images from a real dataset, we observed a significant presence of outliers in the depth map. Those outliers are undetected points to which the sensor assigns $0$ depth. To increase robustness against outliers, the loss function was modified to mask out the extremes of the normalized depth interval, thus avoiding the undetected points to be decoded as close points.
\subsubsection*{Results}
Results are presented in Fig.~\ref{fig:cabinet_exp_synthetic_scene}-\ref{fig:cabinet_exp_real_conditional_likelihoods}. We evaluate the CAE on two depth input images, acquired from the synthetic data generator described in Sec.~\ref{sec:implementation_training} and from a real dataset, respectively. The loss function $\mathcal{L}$ is plotted as a function of the considered four-dimensional state. In the synthetic test-case, the CAE model gives comparable results as the \textit{Synthetic\&Synthetic} and the \textit{Input\&Synthetic} benchmarks. The loss function plots of Fig. ~\ref{fig:cabinet_exp_synthetic_conditional_likelihoods} all have a minimum at the ground truth state, which shows that the network succeeds in learning the dependency on the state. This is in accordance with the generated image of  Fig.~\ref{fig:cabinet_exp_synthetic_reconstruction}, where a new depth image is reconstructed at a specified door angle.

Evaluation for the real test scenario (last row of Fig.~\ref{fig:likelihood_plots}) is more challenging due to the considerable measurement noise in the input depth image, shown in Fig.~\ref{fig:cabinet_exp_real_input_image}.
The CAE model proves to be robust against the measurement noise and, as for the synthetic test, is able to generate different images at specified door angles.
%On the other hand, the \textit{Input\&Synthetic} benchmark fails to provide acceptable results along the $y$ translation, yaw rotation, and door opening angle.
%This is motivated by the presence of holes in the input depth image, as well as regions corresponding to objects that cannot be recreated by the synthetic data generator.
The performance of the learned model in estimating the likelihood is on par with the \textit{Synthetic\&Synthetic} benchmark, the upper bound of what can be achieved with the proposed method. It outperforms the \textit{Input\&Synthetic} benchmark, which fails for translations in $y$ direction and rotations around $z$ due to self occlusions.
\section{CONCLUSIONS \& FUTURE WORK}
\label{sec:conclusions}
DMU has shown to be a general framework to derive measurement update rules for Bayes filters. The learned models outperform the synthetic data baseline solution in the accuracy of the computed likelihood. Even though the models are only trained on synthetic data, they perform well on real-world data, too.
Compared to A-SDF, we report a 1000x inference speedup in articulation state estimation, making the deployment in robotic applications feasible.

Furthermore, a particle filter using DMU offers interesting perspectives for active perception. In a scenario such as the example shown in Fig.~\ref{fig:pf2_color} where the particle filter proposes two possible regions where a target object could be hidden, a robot can be commanded to explore those regions actively.
%In the presented integration experiment, we showed how the DMU network can be used to compute the observation likelihood of a given set of particles in forward mode. Since the trained model is differentiable, gradients with respect to the state vector can easily be computed.
%This allows for the deployment of DMU in an Extended Kalman Filter.
%The gradients can also be used to directly infer the system state by minimizing the loss via gradient descent on a single measurement or as part of a batch-optimization-based state estimator on a series of input images.

Solely providing depth information, the learned models have to rely on geometric properties to distinguish the modeled and unmodeled aspects of the observed system. We identified this as the major limitation during the inspection of failure cases on the RBO dataset.
In future work, we would like to use RGB images as well, since texture greatly simplifies the detection of important features in complex scenes.
Extending this work to RGB-D images poses additional challenges, because the unmodeled state dimension is much larger than for depth-only, increasing the imbalance between background and state dimensionality.
Instead, we propose to use segmentation masks as an additional input since they are easy to obtain from the simulator for training and from a separate instance segmentation network for deployment. 
%In addition, it is much harder to generate realistic RGB images in simulation, which would facilitate the sim-to-real transfer.
% Unfortunately, the proposed method does not trivially generalize to RGB-D inputs since the latent vector compression is not expressive enough for colored scenes. A potential angle of attack could be the use of a U-Net \cite{unet2015} like encoder-decoder structure with skip-connections at different compression levels. Another possible solution could be a multi-stage approach in which a RGB segmentation network computes segmentation masks that serve as an additional input layer to the DMU.

\bibliographystyle{IEEEtran}
\bibliography{dmu}
\end{document}